\documentclass{article} 
\usepackage{iclr2026_conference,times}

\usepackage{amsmath,amsfonts,bm}

\def\eqref#1{equation~\ref{#1}}

\def\1{\bm{1}}

\DeclareMathAlphabet{\mathsfit}{\encodingdefault}{\sfdefault}{m}{sl}
\SetMathAlphabet{\mathsfit}{bold}{\encodingdefault}{\sfdefault}{bx}{n}

\usepackage{url}
\usepackage{algorithm}
\usepackage{algorithmic}
\usepackage{multirow}
\usepackage{amsmath}
\usepackage{amssymb}
\usepackage{tikz}
\usepackage{pgfplots}
\pgfplotsset{compat=1.18} 
\usepackage{pgf-pie}
\usepgfplotslibrary{statistics}
\usepackage{booktabs}
\usepackage{graphicx}
\usepackage{amsmath}
\usepackage{amsfonts}
\usepackage{multirow}
\usepackage{graphicx}
\usepackage{wrapfig}
\usepackage{colortbl}
\usepackage{tikz}
\usepackage{pgfplots}
\pgfplotsset{compat=1.18}
\usepackage{booktabs}
\usepackage{pdflscape}  
\usepackage{adjustbox}  
\usepackage{kotex}
\usepackage{afterpage}
\usepackage{pgf-pie} 
\usepackage{hyperref}
\usepackage{cleveref}
\usepackage{xspace}
\usepackage{xcolor}
\usepackage{lineno}
\usepackage{cuted}
\usepackage{dblfloatfix}
\usepgfplotslibrary{statistics}
\newcommand{\ie}{i.e.\@\xspace}

\title{MSG Score: Automated Video Verification for Reliable Multi-Scene Generation}

\author{
Daewon Yoon$^{1,2*}$, Hyeongseok Lee$^{1*}$, Wonsik Shin$^{1}$, Sangyu Han$^{1}$, Nojun Kwak$^{1\dagger}$ \\
$^{1}$Seoul National University $^{2}$Samsung Electronics \\
$^{*}$ Equal contribution $^{\dagger}$ Corresponding author
}

\iclrfinalcopy 
\begin{document}

\maketitle


\begin{abstract}
While text-to-video diffusion models have advanced significantly, creating coherent long-form content remains unreliable due to stochastic sampling artifacts. This necessitates generating multiple candidates, yet verifying them creates a severe bottleneck; manual review is unscalable, and existing automated metrics lack the adaptability and speed required for runtime monitoring. Another critical issue is the trade-off between evaluation quality and run-time performance: metrics that best capture human-like judgment are often too slow to support iterative generation. These challenges, originating from the lack of an effective evaluation, motivate our work toward a novel solution.

To address this, we propose a scalable automated verification framework for long-form video. First, we introduce \textbf{the MSG(Multi-Scene Generation) score}, a hierarchical attention-based metric that adaptively evaluates narrative and visual consistency. This serves as the core verifier within our \textbf{CGS (Candidate Generation and Selection) framework}, which automatically identifies and filters high-quality outputs. Furthermore, we introduce \textbf{Implicit Insight Distillation (IID)} to resolve the trade-off between evaluation reliability and inference speed, distilling complex metric insights into a lightweight student model. Our approach offers the first comprehensive solution for reliable and scalable long-form video production.
\end{abstract}   
\section{Introduction}

In recent years, diffusion models have advanced from high-fidelity still image generation~\cite{Ho:20, Nichol:21} to addressing the complex challenges of video generation~\cite{Singer:22}. However, this progress has exposed critical issues for film-level content. We identify three fundamental bottlenecks. 

\begin{figure*}[htp!]
\centering
\includegraphics[width=\textwidth]{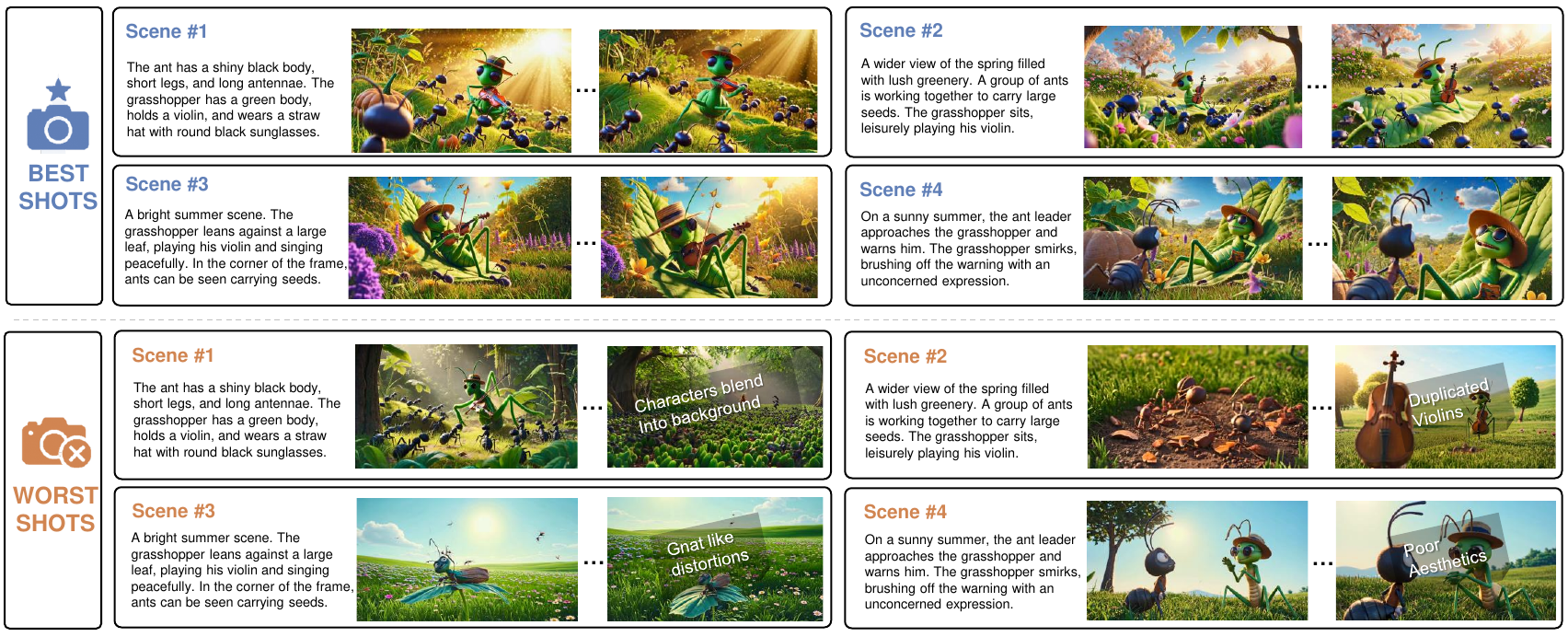}
\caption{Example of long-form multi-scene video generation based on the classic Ant and Grasshopper tale story. Using our CGS Framework, diverse outputs were sampled for each prompt and ranked using the proposed MSG score. Selected best shots demonstrate high consistency and fidelity, while discarded worst cases exhibit background-character blending, object duplication, temporal distortion, and low aesthetic quality.}
\label{fig:figure1}
\end{figure*}

First, there is no truly unified video evaluation score. While comprehensive suites like VBench~\cite{Huang:23} and EvalCrafter~\cite{Liu:24} organize dozens of sub-metrics, they lack a unified score for adaptive, film-level results, which are composed of scenes formed by individual 5-second shots. The importance of each metric shifts depending on a video's scene. For instance, action scenes prioritize motion quality, while portraits prioritize character consistency, forcing creators to manually weigh these conflicting signals. Second, existing evaluation methods such as DOVER~\cite{Wu:23:DOVER} and AIGV Assessor~\cite{Wang:24b} are shot-centric and fail to account for inter-shot consistency, which is a crucial element for creating coherent, long-form narratives from multiple 5-7 second clips. Finally, applying this fragmented and manual evaluation process to the numerous candidates generated for a multi-scene video is prohibitively labor-intensive and slow. To address this, we introduce a comprehensive solution built on three synergistic contributions.

First, we introduce the MSG(Multi-Scene Generation) score, a core evaluation engine designed to solve the problem of scene-agnostic metrics by intergrating them into a single score. It consists of two components: a Perceptual Quality Score and a Thematic Consistency Score. The MSG score achieves an accuracy of 93.48\% in aligning with human judgments. However, a score alone is insufficient. To replace the brute-force generation and inefficient manual review process, we introduce the CGS(Candidate Generation and Selection) framework. This end-to-end system uses the MSG score to efficiently manage the production pipeline, from generating diverse candidates to automatically selecting the best shots and enabling feedback-driven regeneration. Finally, to make this framework practical at scale, the IID(Implicit Insight Distillation) method resolves the critical trade-off between evaluation quality and speed. By distilling the knowledge of our comprehensive MSG score into a lightweight student model, IID achieves a dramatic 61.6x speed-up in evaluation time at the cost of only a 10.1\% decrease in performance. Our contributions can be summarized as follows:

\begin{itemize}
\item We introduce the MSG score, a comprehensive and unified benchmarking. It combines perceptual video quality and thematic consistency. The perceptual quality model employs temporal and spatial features to interpret contextual evaluation. In addition, thematic consistency model ensures coherent character or object between multi-scenes. The MSG score closely matches human judgments, with an accuracy of 93.48\%.
\item We propose the CGS framework, a methodological pipeline that replaces brute-force generation and labor-intensive manual review with an efficient and scalable workflow. Figure~\ref{fig:figure1} shows the effectiveness of our framework in distinguishing between high-quality and flawed candidate scenes.
\item We present the IID method, a teacher-student approach that resolves the trade-off between evaluation quality and runtime performance, enabling fast yet highly accurate automated assessment. This makes high performance by 61.6x speed-up, automated assessment truly feasible for real-world, iterative workflows. 
\end{itemize}

\section{Related Works}
\subsection{Video Generation}
Diffusion models have rapidly advanced video synthesis, building on the foundational image generation work~\cite{Ho:20}. Significant research, including models such as Make-A-Video~\cite{Singer:22}, and Imagen Video~\cite{Saharia:22}, has focused on improving the fidelity, resolution, and temporal consistency of single video clips generated from text. However, a challenge common to all these state-of-the-art generative models is their probabilistic nature. Achieving a desired outcome often requires generating numerous candidates and engaging in a time-consuming, trial-and-error process of manual selection. This "selection bottleneck" severely hinders the practical application and scalability of these powerful models. Our work addresses this universal problem not by proposing another generative model, but by introducing the MSG score and the CGS framework designed to evaluate and automate this critical process for the best long-form video.

\subsection{Video Generation Evaluation}
The evaluation of generated video has matured from simple extensions of image metrics to more sophisticated. Early methods focused on distributional similarity (e.g., FVD~\cite{unterthiner2019fvd}). More recently, comprehensive benchmark suites such as VBench~\cite{Huang:23}, EvalCrafter~\cite{Liu:24}, DOVER~\cite{Wu:23:DOVER}, and AIGV-Assessor~\cite{Wang:24b} have emerged like Appendix A, offering multidimensional analysis across dozens of quality attributes. While these tools excel at offline analysis, they are computationally too intensive to serve as real-time feedback signals for automated verification. They lack the speed required for iterative ranking and selection loops. Our MSG score addresses this critical gap, providing a unified and scalable evaluation engine—optimized via distillation—specifically designed for automated high-quality video production.
\section{Method}

Our approach to scalable long-form video generation is built upon three core technical contributions. We begin by detailing \textbf{the MSG score}, a unified metric that assesses videos on two fronts: perceptual quality, evaluated by a hierarchical attention model over spatial and temporal features, and thematic consistency, which ensures narrative cohesion across scenes. This comprehensive score then powers \textbf{the CGS framework}, an automated pipeline that not only ranks and selects candidates but also utilizes a feedback loop to refine prompts when generation quality is suboptimal. Finally, to ensure the entire framework operates at a practical speed, we introduce \textbf{the IID method}, which resolves the critical speed-accuracy trade-off by using teacher-student distillation to create an efficient yet powerful evaluator.

\subsection{MSG(Multi-scene Generation) Score}
\textbf{The MSG Score} is a unified metric that consolidates attention-derived perceptual quality and VQA-based thematic consistency through a hierarchical weighting mechanism.

\subsubsection{Perceptual Quality Score}
Given a shot $S$, we first identify a set of $m = 20$ widely applicable metrics from recent video evaluation methods, capturing aspects like picture fidelity, prompt adherence, and aesthetic style.


Since video evaluation requires understanding both spatial and temporal contexts, we employ a hierarchical attention structure. First, a \textit{spatial attention layer} processes the feature vector of each shot to model the interdependencies between different metrics within a single frame of reference. Subsequently, a \textit{temporal attention layer} processes the sequence of these representations across multiple shots to capture their evolution and dynamics. This process yields a context-aware feature vector, $H_{\text{context}} \in \mathbb{R}^{d}$, for each shot, which is then fed into a final fusion layer. The \textit{fusion attention layer} computes the final perceptual quality score by applying an attention mechanism to the enriched representation, $H_{\text{context}}$. Specifically, we compute query, key, and value vectors using learnable parameters $W_Q, W_K, W_V \in \mathbb{R}^{d \times d}$:
\begin{equation}
\begin{split}
H_{\text{fused}} &= \mathrm{Att}\bigl(W_Q H_{\text{context}}, W_K H_{\text{context}}, W_V H_{\text{context}}\bigr) \\
&= (W_Q H_{\text{context}} H_{\text{context}}^T W_K^T/\sqrt{d}) W_V H_{\text{context}}
\end{split}
\end{equation}

\begin{figure*}[!tb]
\centering
\includegraphics[width=\textwidth]{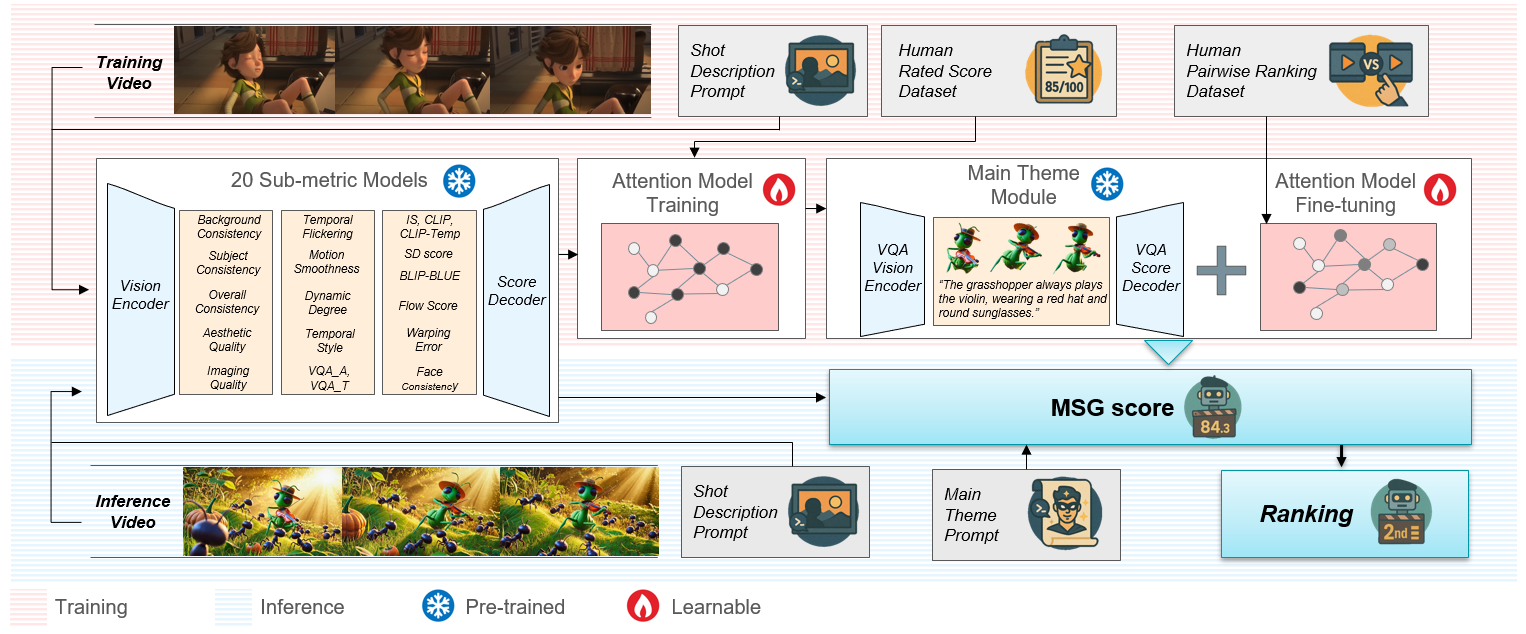}
\caption{Training and inference pipeline of the MSG Score. We construct a 20-dimensional input feature by selecting reliable submetrics(pre-trained) from VBench and EvalCrafter. The attention-based scoring model is trained on quality-labeled data from VBench and AIGV-Assessor, and fine-tuned with pairwise rankings to improve accuracy. During inference, MSG Score is computed by adaptively weighting the submetrics. Meanwhile, the Main Theme module segments the main character and evaluates consistency with the prompt using a pre-trained VQA model.}
\label{fig:figure2}
\end{figure*}

A final projection $W_O \in \mathbb{R}^{1\times d}$ is applied, followed by a GELU non-linearity. The entire process can be conceptualized as applying a dynamic weight matrix $W(S)$ to the contextualized features to produce the perceptual quality score, $\text{MSG}_{\text{quality}}(S)$ where the dynamically computed weight vector $W(S)$ has dimensions $1 \times d$. 
\begin{equation}
\begin{split}
o(S) &= W_O \cdot H_{\text{fused}} \triangleq W(S) H_{\text{context}} \\
\text{MSG}_{\text{quality}}(S) &= \mathrm{GELU}(o(S))
\end{split}
\end{equation}
Note that our weighting mechanism is shot-dependent, \ie the weight matrix $W$ varies with the shot $S$, unlike a simple multi-layer perceptron (MLP) which produces shot-independent weights.

\subsubsection{Thematic Consistency Score}
In parallel, the Main Theme module is dedicated to preserving global creative cohesion. It achieves this by treating the prompt as a guiding theme, encoding it into a reference feature $g(\text{prompt})$ and each shot into a feature $g(S_i)$ to measure key attributes like character or object.

For this cross-modal encoding task, we employ a VQA-based method. Unlike general-purpose Vision Transformers (ViTs), which can lose fine-grained spatial information and are not inherently optimized for text-image alignment, VQA-based models are suit for such reasoning. This specialization provides a more direct and reliable measure of thematic consistency.

Specifically, we measure the overall discrepancy as follows:
\begin{equation}
\mathcal{S}_{\text{global}} \;=\; \frac{1}{N}\sum_{i=1}^{N} \Bigl\|\; g(S_i) \;-\; g(\text{prompt})\Bigr\|_2^2
\end{equation}
This discrepancy is then transformed into a scalar consistency score $z$ (e.g., $z = \exp(-\mathcal{S}_{\text{global}})$), where a higher value indicates stronger thematic adherence.

\subsubsection{Score Integration and Training}
Finally, the comprehensive MSG score for a shot $S$, denoted as $\text{MSG}_{\text{final}}(S)$, is computed by combining the perceptual quality score with the global consistency score $z$. With $\lambda$ empirically set to 0.2, the final score is:
\begin{equation}
\text{MSG}_{\text{final}}(S) \;=\; (1-\lambda)\text{MSG}_{\text{quality}}(S) \;+\; \lambda z
\end{equation}

During training, the learnable parameters of our model ($W_Q, W_K, W_V, W_O$) are jointly optimized to align its predictions with human quality judgments, as illustrated in Figure~\ref{fig:figure2}. To this end, we utilized video samples from several established benchmarks proposed in related work, including VBench~\cite{Huang:23}, EvalCrafter~\cite{Liu:24}, and AIGV-Assessor~\cite{Wang:24b}. These benchmarks provide a diverse range of both high-quality and degraded examples (e.g., prompt mismatches, flickering, motion blur, and temporal inconsistency) reflecting real-world conditions. Additionally, to address the relative scarcity of flawless videos in these benchmarks, we curated a small set of near-perfect samples to better represent the upper bound of perceived quality. The training objective is to minimize the Mean Squared Error (MSE) loss between the predicted perceptual score of the model, $\text{MSG}_{\text{quality}}$ and the human annotations, $y_j$:

\begin{equation}
\mathcal{L}_{\text{MSE}} = \frac{1}{N}\sum_{j=1}^{N}\bigl(\text{MSG}_{\text{quality}}(S_j) - y_j\bigr)^2
\end{equation}

Here, \(y_j\) represents the human-annotated quality label for the \(j\)-th shot, and \(N\) is the total number of training samples.

\subsection{CGS(Candidate Generation and Selection) Framework}
The CGS framework operationalizes the MSG score to create a fully automated, end-to-end pipeline for multi-scene video production. It replaces the labor-intensive cycle of manual generation and review with an systematic workflow. The framework operates in three stages:

\begin{enumerate}
    \item \textbf{Candidate Generation:} For each prompt describing a scene in the narrative, a text-to-video generation model produces a diverse set of candidate shots.

    \item \textbf{Automated Evaluation and Ranking:} Each generated candidate is then automatically evaluated using our comprehensive $\text{MSG}_{\text{final}}$ score. The framework ranks all candidates for a given prompt based on this score.

    \item \textbf{Selection and Feedback Loop:} The top-ranked candidate for each prompt is selected to be part of the final video. Crucially, the framework includes a quality control mechanism. If the highest score among all candidates for a prompt falls below a predefined threshold $T$, the generation is flagged as a failure. This can trigger an automated feedback loop that can refine the prompt and initiate a new generation cycle.
\end{enumerate}

\subsection{IID (Implicit Insight Distillation) Method}
To resolve the runtime bottleneck caused by calculating 20 sub-metrics, we introduce the IID method via knowledge distillation. The full MSG model acts as the \textbf{Teacher}, providing accurate but computationally expensive soft labels. The \textbf{Student} is a lightweight model trained to predict these scores directly from visual patches, effectively bypassing the slow feature extraction process. The student minimizes the distillation loss $\mathcal{L}_{\text{IID}} = \mathbb{E}_{S \sim \mathcal{D}} [ (M_{\text{student}}(S) - M_{\text{teacher}}(S))^2 ]$, inheriting the teacher's nuanced insights. This distillation achieves dramatic speed-ups while maintaining accuracy, making the framework scalable for real-world automated production.

\section{Experiments}
\label{sec:experiments}
To validate our proposed method, we conducted a series of experiments designed to answer three key research questions: (1) Does our CGS framework, powered by the MSG Score, provide a tangible efficiency compared to manual methods? (2) Does the MSG Score's internal architecture, particularly the Main Theme module and its VQA-based encoder, make robust and justifiable decisions? (3) How does the unified MSG Score interpret the trade-offs between various low-level quality metrics? All experiments were conducted on a single NVIDIA A6000 GPU. We are also publicly releasing our evaluation dataset, which includes a diverse set of videos, to encourage further research.

\subsection{Efficiency Compared to Manual Methods}
\paragraph{Workflow Efficiency}
To analyze workflow efficiency, we timed the creation of a 1-minute video (12 five-second shots), selecting the best from 5 candidates per shot. The Manual Workflow required 300 seconds (12 shots × 5 candidates × 5s) of pure playback time. In contrast, our MSG-Assisted Workflow eliminates this review time entirely by automatically selecting the top-ranked candidate.

\paragraph{IID Evaluation Performance}

The experiment was designed to measure the time required to evaluate 300 generated video clips of 5-second length. We compared the heavyweight teacher model (the full MSG Score model), which computes all 20 sub-metrics, against the lightweight student model, which is distilled via IID. The results clearly demonstrated a dramatic performance improvement with the IID method.

The teacher model took a total of \textbf{308 minutes} to evaluate 300 videos, whereas the student model using the IID method completed the same task in just \textbf{5 minutes}. This resulted in a \textbf{61.6x speed-up} in evaluation time. Remarkably, this dramatic gain in efficiency came at the cost of only a \textbf{10.01\%} decrease in performance, as measured by the average error rate. Detailed per-feature results are presented in a table in the Appendix A. This crucial improvement, reducing a multi-hour task to minutes with only a 10.01\% performance decrease, makes our CGS framework a truly practical and scalable solution.

\begin{figure}[!bp]
    \centering
    \begin{tikzpicture}
    \begin{axis}[
        width=9cm,
        height=7cm,
        ymin=0, ymax=5.5,
        ylabel={5-point Likert User Rating},
        ylabel style={yshift=-0.2em},
        xtick={1,2,3,4},
        xticklabels={
            {\shortstack{Cinematic\\high MSG}},
            {\shortstack{Cinematic\\low MSG}},
            {\shortstack{Animated\\high MSG}},
            {\shortstack{Animated\\low MSG}}
        },
        boxplot/draw direction=y,
        boxplot={box extend=0.4},
        boxplot/.style={semi thick},
        whisker/.style={semi thick},
        median/.style={ultra thick},
    ]

    \addplot+[
        boxplot prepared={
            lower whisker=3,
            lower quartile=4.0,
            median=5.0,
            upper quartile=5.0,
            upper whisker=5
        },
        boxplot/draw position=1,
        style={fill=gray!20, draw=gray},
        mark=none
    ] coordinates {(0,4.37)};
     \node at (axis cs:1.35,4.37) {\small \textbf{\textcolor{black}{4.37}}};

    \addplot+[
        boxplot prepared={
            lower whisker=1,
            lower quartile=1.0,
            median=2.0,
            upper quartile=3.0,
            upper whisker=5
        },
        boxplot/draw position=2,
        style={fill=gray!20, draw=gray},
        mark=none
    ] coordinates {(0,2.05)};
    \node at (axis cs:2.35,2.05) {\small \textbf{\textcolor{gray}{2.05}}};

    \addplot+[
        boxplot prepared={
            lower whisker=3,
            lower quartile=4.0,
            median=5.0,
            upper quartile=5.0,
            upper whisker=5
        },
        boxplot/draw position=3,
        style={fill=gray!20, draw=gray},
        mark=none
    ] coordinates {(0,4.47)};
     \node at (axis cs:3.35,4.47) {\small 
   \textbf{ \textcolor{gray}{4.47}}};

    \addplot+[
        boxplot prepared={
            lower whisker=1,
            lower quartile=1.0,
            median=2.0,
            upper quartile=3.0,
            upper whisker=5
        },
        boxplot/draw position=4,
        style={fill=gray!20, draw=gray},
        mark=none
    ] coordinates {(0,2.02)};
    \node at (axis cs:4.35,2.02) {\small 
    \textbf{\textcolor{gray}{2.02}}};
    \addplot+[
    only marks,
    mark=*,
    color=blue
    ] coordinates {
        (1, 4.37)
        (2, 2.05)
        (3, 4.47)
        (4, 2.02)
    };
    \node at (axis cs:2.5,0.5) {\small \textit{\textcolor{blue}{Mean values are represented by blue points}}};

    \end{axis}
    \end{tikzpicture}
    \caption{
       Survey results based on a 5-point Likert scale summarize user ratings. The box plot presents mean values and IQR-based statistics, illustrating a strong alignment between MSG Score rankings and human perception.
    }
    \label{fig:stick}
\end{figure}

\subsection{Robustness of the MSG Score}

\paragraph{Alignment with Human Perception}
To demonstrate the comprehensive usefulness of our approach, we conducted two user studies via an online survey involving 43 participants (16 general users, 16 AI professionals, and 11 film professionals).

The first study assessed if a unified score could relieve the burden of manual review. We asked participants, \emph{``Assuming you create a long-form video using AI... which quality evaluation method do you prefer?''}. The choices were \emph{(i)} a single, unified score or \emph{(ii)} 16 multiple domain-based metrics. A statistically significant majority of 26 participants (60.5\%) preferred the single integrated score (Mann–Whitney U test: $p = 0.048$, Cliff δ = –0.35), emphasizing its ease of use and potential for faster iteration.

The second study evaluated how closely our MSG Score aligns with subjective human judgment. The same 43 participants rated four representative clips selected based on our score: \textit{Cinematic Good} and \textit{Bad} (from the same prompt), and \textit{Animated Good} and \textit{Bad} (from the same prompt). As summarized in Figure~\ref{fig:stick}, the results show a strong correlation: the two \emph{Good} videos received high average ratings of 4.37 and 4.47, whereas the \emph{Bad} videos were rated at 2.05 and 2.02. This clearly demonstrates the alignment between the MSG Score’s assessments and human perception, reinforcing its validity for real-world usage.

\begin{table}[!tp]
\centering
\caption{Comparison between ViT (DINOv2) and VQA (GPT-4o) on thematic consistency scoring across video frames. The VQA model proves more robust at detecting subtle flaws (Case 1) and maintaining high scores for coherent but degraded scenes (Case 2).}
\label{tab:vit_vqa_comparison}
\begin{tabular}{l cc cc}
\toprule
\multirow{2}{*}{\textbf{Frames}} & \multicolumn{2}{c}{\shortstack{\textbf{Case 1} \\ \small (Subtle Flaws)}} & \multicolumn{2}{c}{\shortstack{\textbf{Case 2} \\ \small (Blurry but Coherent)}} \\
\cmidrule(lr){2-3} \cmidrule(lr){4-5}
& ViT $\uparrow$ & VQA $\uparrow$ & ViT $\uparrow$ & VQA $\uparrow$ \\
\midrule
1 & 0.980 & 0.710 & 0.922 & 0.999 \\
2 & 0.946 & \textbf{0.412} & \textbf{0.893} & 0.999 \\
3 & 0.964 & 0.783 & 0.929 & 0.998 \\
4 & 0.969 & 0.985 & 0.954 & 0.999 \\
\bottomrule
\end{tabular}
\end{table}

\paragraph{Justification for VQA-based Thematic Consistency.}
To empirically validate our choice of a VQA-based model, we compared its performance against a state-of-the-art, image-based encoder (ViT, specifically DINOv2-Giant). We analyzed two challenging cases, with results summarized in Table~\ref{tab:vit_vqa_comparison}. \textbf{Case 1} involved a video with subtle, intermittent visual flaws (e.g., face warping). The ViT model assigned consistently high similarity scores, failing to penalize the flawed frames. In contrast, the VQA model (GPT-4o) correctly identified the inconsistency, with its score dropping significantly on the corrupted frame. \textbf{Case 2} featured a video that was thematically consistent but visually degraded (e.g., blurry). Here, the ViT's scores fluctuated, sensitive to the low-level degradation despite the scene's coherence. The VQA model, however, provided near-perfect scores, correctly recognizing the underlying thematic consistency. These findings confirm that VQA-based models are more robust for assessing high-level semantic coherence than purely visual encoders.

\begin{figure}[!bp]
    \centering
    \caption{Comparison of video quality with and without the Main Theme method. Without the Main Theme, video scores were similar regardless of character appearance. Enabling the method led to a clear divergence: consistent character depiction (Case 1) resulted in higher MSG scores, while deteriorating appearance (Case 2) produced lower scores.}
    \label{fig7}
    \begin{minipage}{0.4\textwidth}
        \centering
        \includegraphics[width=\linewidth]{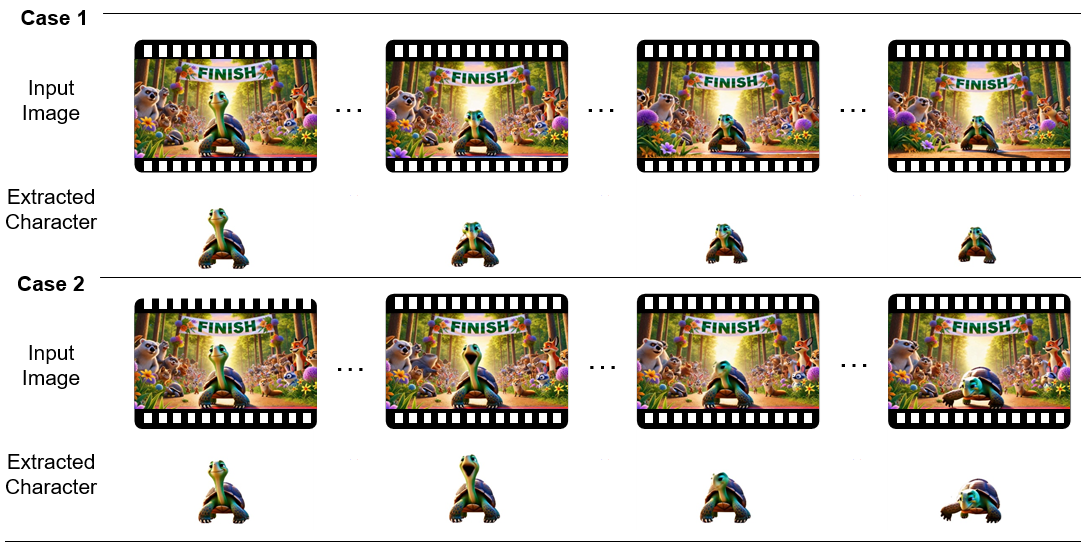}
        \label{fig:video_sample}
    \end{minipage}%
    \hfill
    \begin{minipage}{0.6\textwidth}
        \centering
        \begin{tabular}{lccccc}
            \toprule
            Videos  & MSG Score & MSG Score \\
                    & without Main Theme & with Main Theme \\
            \midrule
            Case 1 & \textcolor{black}{51.68} & \textcolor{blue}{60.34}\\
            Case 2 & \textcolor{black}{52.54} & \textcolor{red}{46.15} \\
            \bottomrule
        \end{tabular}
    \end{minipage}
\end{figure}

\paragraph{Main Theme On vs.\ Off.}
To investigate how robustly our \textit{Main Theme} method enforces global thematic and stylistic coherence, we ran an ablation study toggling Main Theme method on/off. Specifically, we generated the same multi-scene videos, once allowing the main theme mechanism to unify character and stylistic elements, and once without it. Before applying the Main Theme method, the overall video quality scores were nearly same. However, Once we incorporated the Main Theme VQA to quantify how closely each frame’s tortoise matched the appearance defined in the Main Theme prompt, the results changed dramatically as shown in Figure.~\ref{fig7}. In Case 2, where the turtle’s depiction gradually deteriorated in the later frames, the MSG score dropped significantly. By contrast, In Case 1, videos that maintained a consistent appearance of the main character across all frames achieved significantly higher MSG scores.

\paragraph{Generalization to Human-Made Videos}
To test the generalization capabilities of our MSG Score beyond AI-generated content, we also evaluated it on a set of human-made animations. As detailed in the Appendix C, the MSG Score successfully distinguished between professionally produced animations and low-quality amateur animations, assigning them high and low scores respectively. This result demonstrates that our score has learned universal quality attributes and is not limited to evaluating diffusion-based content.

\subsection{Validating Against Preliminary Approaches}
\label{subsec:preliminary}

\paragraph{Quantitative Comparison}
To quantitatively validate the MSG Score's overall performance, we compared it against DOVER~\cite{Wu:23:DOVER}, an unified video evaluation model. On a test set of 1,200 videos with human-annotated quality scores (0-100), we measured the accuracy of each model in predicting the human judgments. Our MSG Score achieved an accuracy of \textbf{93.48\%}, significantly outperforming DOVER's 86.67\%. This result provides strong quantitative evidence that our hierarchical, adaptive approach aligns more closely with human perception than existing methods.

\begin{figure}[!tp]
    \centering
    \begin{minipage}{0.4\textwidth}
        \centering
        \includegraphics[width=\linewidth]{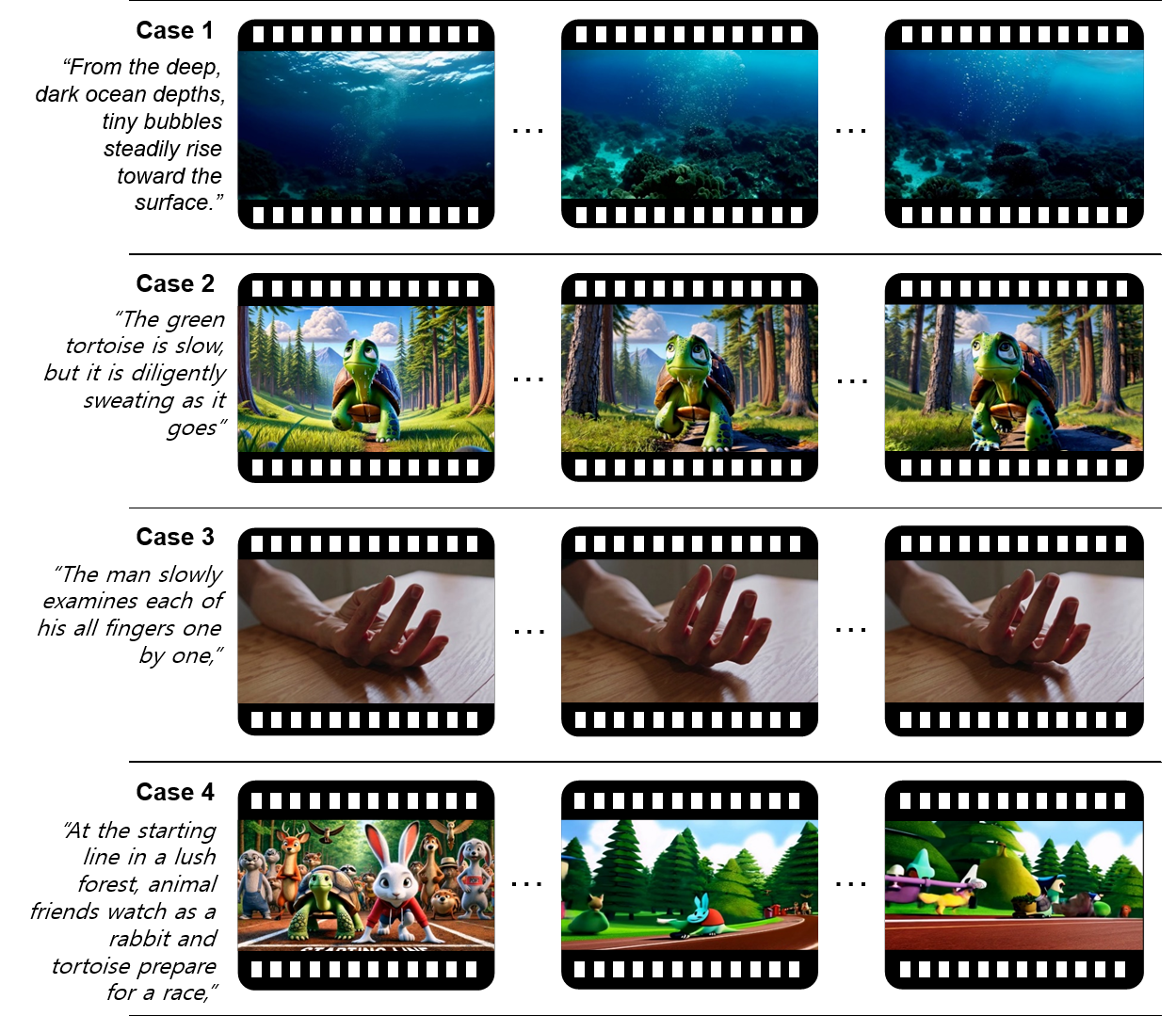}
        \label{fig:video_sample2}
    \end{minipage}%
    \hfill
    \begin{minipage}{0.6\textwidth}
        \centering\begin{tabular}{lcccc}
\toprule
Videos & \multicolumn{3}{c}{\textbf{VBench Suite}} & \multirow{2}{*}{\shortstack{\textbf{MSG}\\\textbf{Score}}}
\\
       & Subject & Aesthetic & Temporal & \\
       & Consistency & Quality & Style & \\
\midrule
Case 1 & \textbf{76.27} & 30.59 & 24.57 & \textcolor{blue}{68.96} \\
Case 2 & 42.70 & 65.74 & 68.78 & \textcolor{blue}{83.97} \\
Case 3 & \textbf{91.53} & \textbf{1.96} & 46.79 & \textcolor{red}{57.09} \\
Case 4 & \textbf{0.71}  & 65.41 & 37.04 & \textcolor{red}{53.81} \\
\midrule
Average & 70 & 55 & 40 & \\
\bottomrule
\end{tabular}
        
    \end{minipage}
\caption{MSG scores for four cases with captured video samples, measured by VBench submetrics and the unified MSG Score. The MSG Score offers a more intuitive overall assessment than viewing submetrics independently.}
\label{fig6}
\end{figure}

\paragraph{Qualitative Comparison}
To gauge the advantage of a unified adaptive MSG score, we compared MSG Score with the individual sub-metrics from VBench~\cite{Huang:23}, such as \emph{Subject Consistency}, \emph{Temporal Style}, \emph{Aesthetic Quality}, etc. Appendix A of this paper overviews these preliminary items. We then examine how these metrics align or conflict with the final MSG Score labels in text-to-video production work. Figure.~\ref{fig6} illustrates four representative cases demonstrating how MSG Score effectively captures nuanced trade-offs among sub-metrics compared to individual VBench measures.

\paragraph{Interesting Cases.}
We observe four distinct behaviors: \textbf{(Case 1)} High scores in critical aspects (e.g., aesthetics) can boost the overall score even if secondary metrics are lower; \textbf{(Case 2)} Balanced, moderate sub-metrics yield a high overall rating; whereas \textbf{(Cases 3 \& 4)} Severe deficiencies in any single dimension significantly penalize the final score, preventing specific failures from being masked by other high attributes. These findings suggest that instead of struggling to interpret the individual prior submetrics, relying on a comprehensive MSG Score offers a more useful and intuitive evaluation of video quality by capturing the nuanced trade-offs among the submetrics.


\section{Conclusion}

In this work, we addressed the critical bottleneck hindering the creation of film-level generative video: the lack of a  video evaluation score and scalable methods. We presented a comprehensive solution built upon three core, synergistic contributions. First, we introduced the MSG Score, a unified and hierarchical metric that provides a nuanced assessment of both perceptual quality and thematic consistency. This score serves as the core engine for the CGS framework, an automated pipeline that replaces laborious manual review with intelligent candidate selection and a feedback-driven workflow. Finally, to make this process practical at scale, we developed the IID method, which uses knowledge distillation to resolve the critical speed-quality trade-off, enabling fast yet highly accurate evaluation.

We demonstrated that the MSG Score achieves 93.48\% accuracy in aligning with human judgments, significantly outperforming a state-of-the-art baseline, DOVER. Moreover, our IID method proved essential for scalability, delivering a 61.6x evaluation speed-up with only a minor drop in performance. The model-agnostic design, validated on both AI-generated and human-made content, establishes our framework as an extensible and general-purpose foundation for future production.

By reframing multi-scene video evaluation as a learning problem, our work lays the groundwork for a paradigm shift from the current "generate-then-filter" model to true "evaluation-aware generation" and interactive content creation. This research not only offers a better tool for evaluation, but lays the foundation for a new paradigm of more intelligent and controllable generative models. Our project page is at \href{https://github.com/anonymous-paper-review/MSGscore}{https://github.com/anonymous-paper-review/MSGscore}.

\clearpage

\bibliography{iclr2026_conference}

@article{Ho:20,
  title={Denoising diffusion probabilistic models},
  author={Ho, Jonathan and Jain, Ajay and Abbeel, Pieter},
  journal={Advances in neural information processing systems},
  volume={33},
  pages={6840--6851},
  year={2020}
}

@inproceedings{Nichol:21,
  title={Improved denoising diffusion probabilistic models},
  author={Nichol, Alexander Quinn and Dhariwal, Prafulla},
  booktitle={International conference on machine learning},
  pages={8162--8171},
  year={2021},
  organization={PMLR}
}

@article{Singer:22,
  title={Make-a-video: Text-to-video generation without text-video data},
  author={Singer, Uriel and Polyak, Adam and Hayes, Thomas and Yin, Xi and An, Jie and Zhang, Songyang and Hu, Qiyuan and Yang, Harry and Ashual, Oron and Gafni, Oran and others},
  journal={arXiv preprint arXiv:2209.14792},
  year={2022}
}

@article{Saharia:22,
  title={Photorealistic text-to-image diffusion models with deep language understanding},
  author={Saharia, Chitwan and Chan, William and Saxena, Saurabh and Li, Lala and Whang, Jay and Denton, Emily L and Ghasemipour, Kamyar and Gontijo Lopes, Raphael and Karagol Ayan, Burcu and Salimans, Tim and others},
  journal={Advances in neural information processing systems},
  volume={35},
  pages={36479--36494},
  year={2022}
}

@article{unterthiner2019fvd,
  title={FVD: A New Metric for Video Generation},
  author={Unterthiner, Thomas and Van Steenkiste, Sjoerd and Kurach, Karol and Marinier, Rapha{\"e}l and Michalski, Marcin and Gelly, Sylvain},
  journal={arXiv preprint arXiv:1812.01717},
  year={2019}
}

@inproceedings{Teed:20,
  title={Raft: Recurrent all-pairs field transforms for optical flow},
  author={Teed, Zachary and Deng, Jia},
  booktitle={European conference on computer vision},
  pages={402--419},
  year={2020},
  organization={Springer}
}

@inproceedings{Radford:21,
  title={Learning transferable visual models from natural language supervision},
  author={Radford, Alec and Kim, Jong Wook and Hallacy, Chris and Ramesh, Aditya and Goh, Gabriel and Agarwal, Sandhini and Sastry, Girish and Askell, Amanda and Mishkin, Pamela and Clark, Jack and others},
  booktitle={International conference on machine learning},
  pages={8748--8763},
  year={2021},
  organization={PmLR}
}

@inproceedings{Huang:23,
  title={Vbench: Comprehensive benchmark suite for video generative models},
  author={Huang, Ziqi and He, Yinan and Yu, Jiashuo and Zhang, Fan and Si, Chenyang and Jiang, Yuming and Zhang, Yuanhan and Wu, Tianxing and Jin, Qingyang and Chanpaisit, Nattapol and others},
  booktitle={Proceedings of the IEEE/CVF Conference on Computer Vision and Pattern Recognition},
  pages={21807--21818},
  year={2024}
}

@inproceedings{Liu:24,
  title={Evalcrafter: Benchmarking and evaluating large video generation models},
  author={Liu, Yaofang and Cun, Xiaodong and Liu, Xuebo and Wang, Xintao and Zhang, Yong and Chen, Haoxin and Liu, Yang and Zeng, Tieyong and Chan, Raymond and Shan, Ying},
  booktitle={Proceedings of the IEEE/CVF Conference on Computer Vision and Pattern Recognition},
  pages={22139--22149},
  year={2024}
}

@inproceedings{Wang:24b,
  title={AIGV-assessor: benchmarking and evaluating the perceptual quality of text-to-video generation with LMM},
  author={Wang, Jiarui and Duan, Huiyu and Zhai, Guangtao and Wang, Juntong and Min, Xiongkuo},
  booktitle={Proceedings of the Computer Vision and Pattern Recognition Conference},
  pages={18869--18880},
  year={2025}
}

@inproceedings{Ke:21,
  title={Musiq: Multi-scale image quality transformer},
  author={Ke, Junjie and Wang, Qifei and Wang, Yilin and Milanfar, Peyman and Yang, Feng},
  booktitle={Proceedings of the IEEE/CVF international conference on computer vision},
  pages={5148--5157},
  year={2021}
}

@article{Kirstain:23,
  title={Pick-a-pic: An open dataset of user preferences for text-to-image generation},
  author={Kirstain, Yuval and Pagra, Uriel and Polyak, Adam and Gat, Uriel and Levi, Tomer and Singer, Yoni},
  journal={arXiv preprint arXiv:2305.01569},
  year={2023}
}

@inproceedings{Li:22:BLIP,
  title={Blip: Bootstrapping language-image pre-training for unified vision-language understanding and generation},
  author={Li, Junnan and Li, Dongxu and Xiong, Caiming and Hoi, Steven},
  booktitle={International conference on machine learning},
  pages={12888--12900},
  year={2022},
  organization={PMLR}
}

@inproceedings{Wu:23:DOVER,
  title={Exploring video quality assessment on user generated contents from aesthetic and technical perspectives},
  author={Wu, Haoning and Zhang, Erli and Liao, Liang and Chen, Chaofeng and Hou, Jingwen and Wang, Annan and Sun, Wenxiu and Yan, Qiong and Lin, Weisi},
  booktitle={Proceedings of the IEEE/CVF International Conference on Computer Vision},
  pages={20144--20154},
  year={2023}
}
\bibliographystyle{iclr2026_conference}

\appendix
\clearpage
\appendix

\newpage

\section*{Appendix A: Detailed Explanation of Evaluation Submetrics}

In this section, we provide detailed definitions and characteristics of the key evaluation submetrics that form the foundation of the MSG Score. These metrics, as detailed in Table~\ref{table1}, are drawn from established benchmarks and assess various aspects of video generation, from low-level visual fidelity to high-level semantic alignment.

\begin{table*}[!b]
  \centering
  \caption{Suite of Evaluation Items and Metrics for VBench, EvalCrafter, and AIGV Assessor}
  \label{table1}
  \small
  \begin{tabular}{lll}
    \toprule
    Benchmark Suite & Evaluation Item & Metric / Method \\
    \midrule
    \multirow{16}{*}{VBench} 
      & Aesthetic Quality        & LAION Aesthetic Predictor score \\
      & Imaging Quality          & MUSIQ image quality score \\
      & Temporal Flickering      & Mean absolute pixel difference across frames \\
      & Motion Smoothness        & Inter-frame interpolation model score \\
      & Dynamic Degree           & RAFT optical flow magnitude \\
      & Background Consistency   & CLIP image feature similarity across frames \\
      & Appearance Style         & CLIP image embedding vs.\ style text similarity \\
      & Temporal Style           & ViCLIP video embedding vs.\ motion style similarity \\
      & Overall Consistency      & ViCLIP video--text alignment score \\
      & Subject Consistency      & DINO feature similarity across frames \\
      & Object Class             & GRiT object detection success rate \\
      & Multiple Objects         & GRiT multi-object detection rate \\
      & Human Action             & UMT action recognition accuracy \\
      & Color                    & GRiT color attribute caption \& match \\
      & Spatial Relationship     & Rule-based spatial relation check \\
      & Scene                    & Tag2Text scene captioning \& match \\
    \midrule
    \multirow{5}{*}{EvalCrafter} 
      & Video Quality            & VQA (aesthetic \& technical scores), Inception Score (IS) \\
      & Text-Video Alignment     & CLIP-Score, BLIP-BLEU, SD-Score \\
      & Content Accuracy         & Detection-Score, Count-Score, Color-Score \\
      & Motion Quality           & Action-Score, Flow-Score, Motion Amplitude (AC-Score) \\
      & Temporal Consistency     & Warping Error, CLIP-Temp, Face ID Consistency \\
    \midrule
    \multirow{4}{*}{AIGV-Assessor} 
      & Static Quality           & Human annotation (clarity, contrast, visual integrity) \\
      & Temporal Smoothness      & Human annotation (frame-to-frame transition smoothness) \\
      & Motion Degree            & Human annotation (strength and diversity of motion) \\
      & Text-Video Correspondence & Human annotation (alignment with input prompt) \\
    \bottomrule
  \end{tabular}
\end{table*}

\paragraph{BLIP-BLEU.} This metric first uses BLIP (Bootstrapped Language Image Pretraining)~\cite{Li:22:BLIP} to generate a textual caption from a video's keyframes. This generated caption is then compared to the original input prompt using the BLEU score. While effective for simple lexical matches, it can underestimate adherence in visually faithful but lexically creative generations.

\paragraph{CLIP-Score.} Measures the cosine similarity between text and image embeddings using CLIP~\cite{Radford:21}. Typically calculated on a per-frame basis, it is robust to paraphrasing but may give high scores even with content-level mismatches if global style features dominate the similarity.

\paragraph{SD-Score.} Based on Stable Diffusion, this score estimates how well a video frame corresponds to a text prompt by reverse-evaluating the conditioning strength. It reflects generative faithfulness but may inherit the biases of the underlying diffusion model.

\paragraph{MUSIQ (Multi-scale Image Quality).} A perceptual quality metric trained on images~\cite{Ke:21}, using multi-scale transformers to consider both global and local patterns. It reflects technical clarity and composition but may penalize intentionally non-photorealistic outputs.

\paragraph{LAION Aesthetic Predictor.} A CLIP-based score trained on crowd-sourced aesthetic preferences~\cite{Kirstain:23}. It provides a scalar value indicating how visually pleasing an image is. However, it can undervalue artistic choices that deviate from conventional beauty.

\paragraph{Temporal Flickering.} Measures the mean absolute pixel difference across consecutive frames to detect high-frequency flicker artifacts, highlighting visual instability. It can, however, be confounded by fast motion or intentional visual effects.

\paragraph{Dynamic Degree.} Computes the average optical flow magnitude using RAFT~\cite{Teed:20} to quantify the extent of motion. While helpful for detecting liveliness, it does not assess whether the motion is semantically meaningful.

\paragraph{Appearance Style.} Evaluates style adherence by comparing CLIP~\cite{Radford:21} embeddings of frames to style-related text descriptions. It captures high-level visual themes but can be sensitive to noise.

\paragraph{Overall Consistency.} Assesses the semantic consistency between the entire video and its prompt. It helps detect global alignment but may overlook local mismatches.

\paragraph{Object Class.} Utilizes object detection models to determine if specific prompt-relevant objects appear. It checks semantic fidelity but may struggle with abstract styles or occlusions.

\paragraph{Multiple Objects.} Based on the same detector, this metric evaluates the presence of multiple, distinct entities as required by the prompt, penalizing under-diversified object compositions.

\paragraph{Color Accuracy.} Measures whether the predicted color attributes of detected objects match the prompt descriptions. Useful for compositional prompts but limited by lighting and shading effects.

\paragraph{Spatial Relationship.} A rule-based check on object positions (e.g., “a cat under a table”). Effective for simple layouts but not robust to visual ambiguity.

\paragraph{Scene Semantics.} Uses scene classification models to label overall scene types and match them to the prompt, helping to confirm high-level setting coherence.

\paragraph{Content Accuracy.} A composite metric that combines detection, count, and color consistency to assess if generated content literally matches the prompt.

\paragraph{Motion Quality.} A set of scores capturing how realistic, continuous, and expressive motion is, helping to detect unnatural transitions or static artifacts.

\paragraph{Temporal Consistency.} Combines multiple techniques such as warping error and identity preservation (e.g., Face ID) to evaluate smoothness and coherence across time.

\paragraph{Static Quality.} A human-labeled metric capturing the clarity, contrast, and visual integrity of individual frames. It provides a perceptual anchor but is expensive to scale.

\paragraph{Temporal Smoothness.} Human-annotated evaluation of the perceptual fluidity of frame-to-frame transitions. It is useful for cinematic realism but subjective.

\paragraph{Motion Degree.} A human-annotated score measuring the strength and variation of motion. Important for action-heavy scenes, it may penalize slow pacing.

\paragraph{Text-Video Correspondence.} A holistic human assessment of how well the video delivers on the semantic and visual expectations set by the prompt. It is the most comprehensive but also the most labor-intensive metric.

\section*{Appendix B: Error Rate by Feature for the Student Model after IID}

As referenced in the main paper, the IID (Implicit Insight Distillation) method significantly accelerates evaluation time. Table~\ref{tab:iid_error_rate_compact} details the per-feature error rates of the lightweight student model when compared to the comprehensive teacher model, demonstrating the effectiveness of the distillation process.

\begin{table}[!ht]
\centering
\caption{Error Rate for the Student Model after IID}
\label{tab:iid_error_rate_compact}
\begin{tabular}{cr cr}
\toprule
\textbf{Feature} & \textbf{Error Rate (\%)} & \textbf{Feature} & \textbf{Error Rate (\%)} \\
\midrule
1 & 5.6 & 11 & 0.5 \\
2 & 2.5 & 12 & 2.3 \\
3 & 1.5 & 13 & 40.2 \\
4 & 1.2 & 14 & 36.9 \\
5 & 1.0 & 15 & 17.4 \\
6 & 9.2 & 16 & 3.7 \\
7 & 6.1 & 17 & 0.2 \\
8 & 6.4 & 18 & 16.2 \\
9 & 5.3 & 19 & 24.2 \\
10 & 9.8 & & \\
\bottomrule
\end{tabular}
\end{table}



\section*{Appendix C: Human-made Video Evaluation by MSG}
\label{appendix:human_video_evaluation}

To assess whether the MSG Score—originally developed for evaluating text-to-video diffusion outputs—can generalize to human-made content, we conducted an evaluation using real-world, manually produced videos with varying levels of production quality. This test is designed to validate whether MSG can offer meaningful discriminative assessments beyond synthetic video generation tasks.

For this experiment, we selected two categories of human-made video data. The first category consisted of professionally produced commercial trailers, specifically including a high-quality promotional animation video from the "Tomb Raider" franchise that was publicly available online. This trailer served as an example of expert-level production, exhibiting strong control over composition, visual effects, and narrative flow. The second category included short video submissions created by middle school students as part of a digital storytelling competition. 

To ensure a fair and uniform evaluation, we assumed that the prompt—or intended narrative—for each video was perfect and fully aligned with the visual content, since it was created by the authors themselves. The MSG Score was then computed for each shot using the same sub-metrics and attention-based weighting model, without any changes to the scoring pipeline.

As shown in Table~\ref{table2}, the results clearly indicate that expert-made videos achieved consistently higher MSG scores (highlighted in red), while the student-created videos generally received lower scores (highlighted in blue). Notably, the MSG Score was particularly sensitive to global inconsistencies, flicker, and prompt alignment issues that were often present in junior-level submissions but absent in the professional trailers.

These findings suggest two important takeaways. First, MSG Score retains its discriminative power even when applied to human-made content, identifying quality differences in a way that aligns with human expectations. Second, the scoring framework does not rely on artifacts specific to diffusion-based generation, but instead evaluates core aspects of visual storytelling and coherence. In conclusion, all shots from expert-level productions surpassed the MSG scores of junior-level productions, confirming the broader generalizability and robustness of our proposed scoring system in evaluating both synthetic and manually crafted long-form videos. It is particularly noteworthy that the MSG Score’s internal attention mechanism successfully differentiated between 'static quality' and 'temporal reliability.' While individual frames in some novice-level submissions exhibited reasonable aesthetics, the temporal attention layers assigned significantly lower weights to sequences with erratic camera motion or abrupt cuts. This confirms that the metric does not merely average frame quality but actively verifies the spatio-temporal continuity essential for high-fidelity video generation.

\begin{table*}[!ht]
\centering
\caption{MSG Scores of Human-made Animation Sets}
\label{table2}

\begin{tabular}{cc}
\multicolumn{1}{c}{\textbf{Expert-made Videos}} & \multicolumn{1}{c}{\textbf{Junior-made Videos}} \\[2mm]
\begin{tabular}{@{}c c@{}}
\toprule
\textbf{Video} & \textbf{MSG Score} \\
\midrule
\includegraphics[height=1.5cm,width=0.15\textwidth]{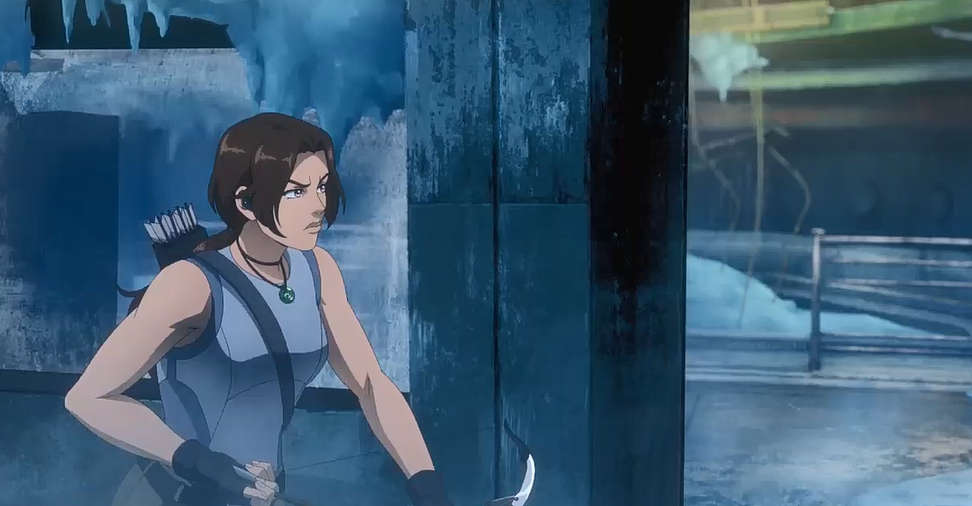}\includegraphics[height=1.5cm,width=0.15\textwidth]{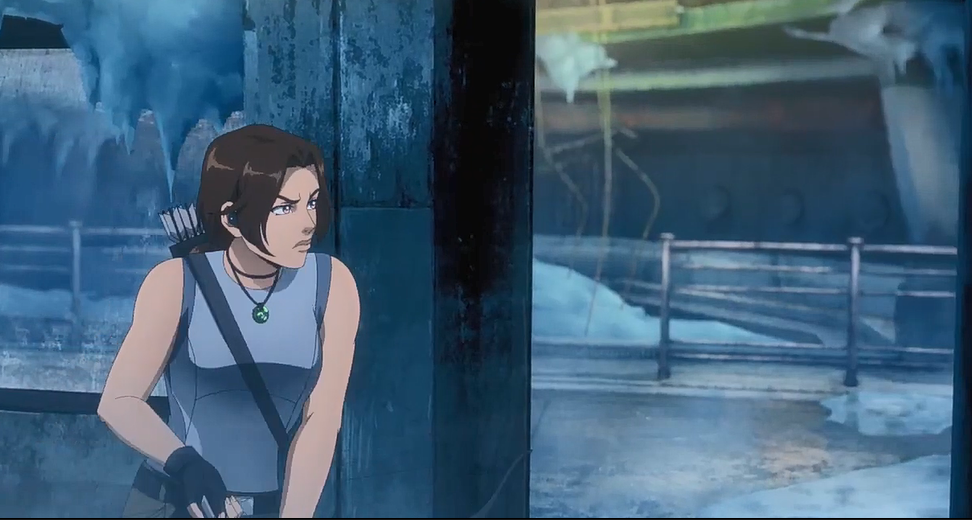} & 39.3583 \\
\includegraphics[height=1.5cm,width=0.15\textwidth]{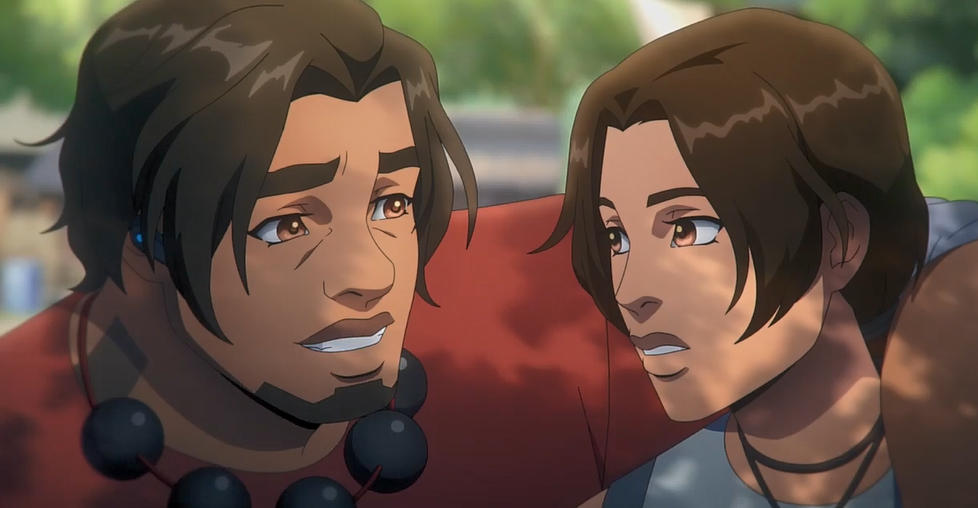}\includegraphics[height=1.5cm,width=0.15\textwidth]{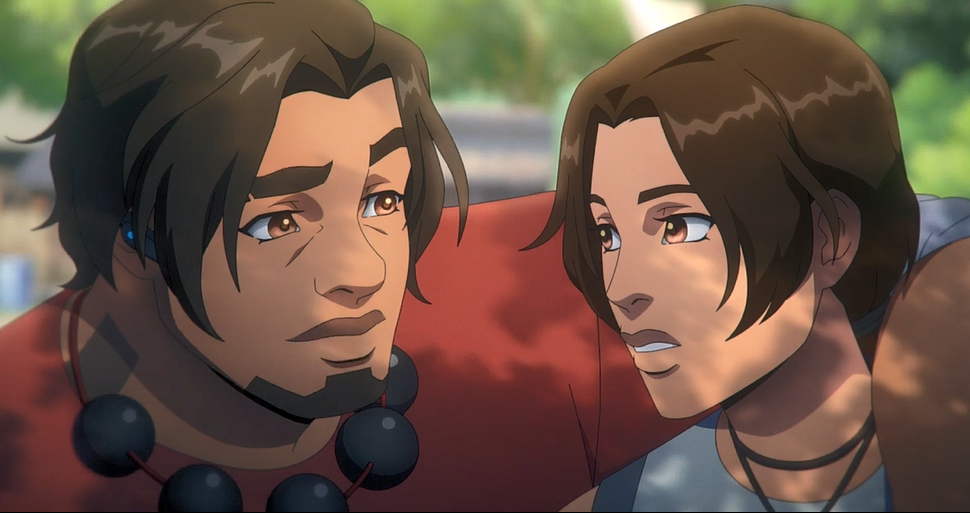} & 42.3005 \\
\includegraphics[height=1.5cm,width=0.15\textwidth]{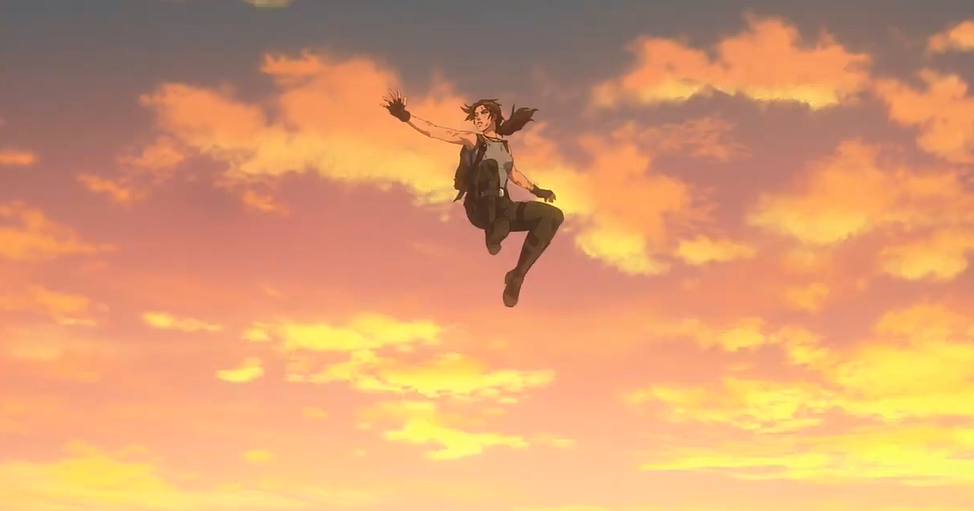}\includegraphics[height=1.5cm,width=0.15\textwidth]{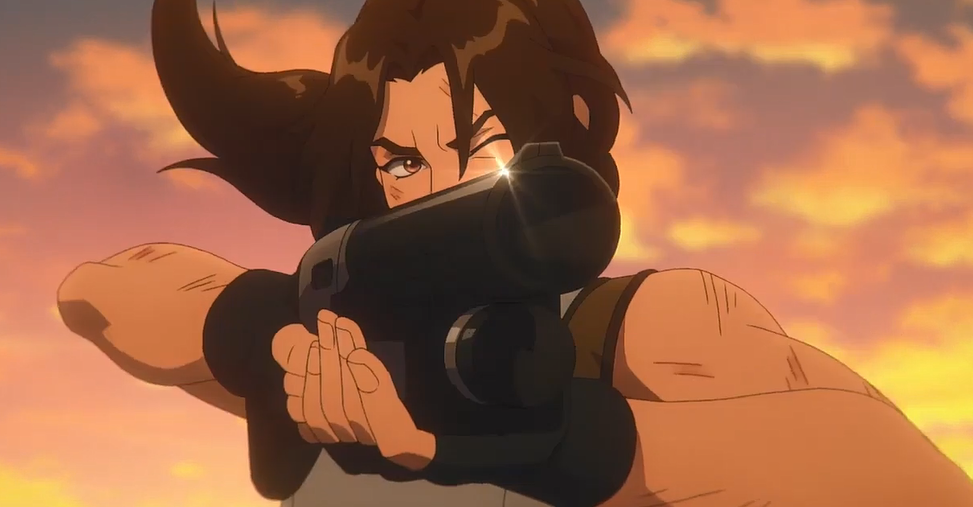} & \textcolor{red}{47.8929} \\
\includegraphics[height=1.5cm,width=0.15\textwidth]{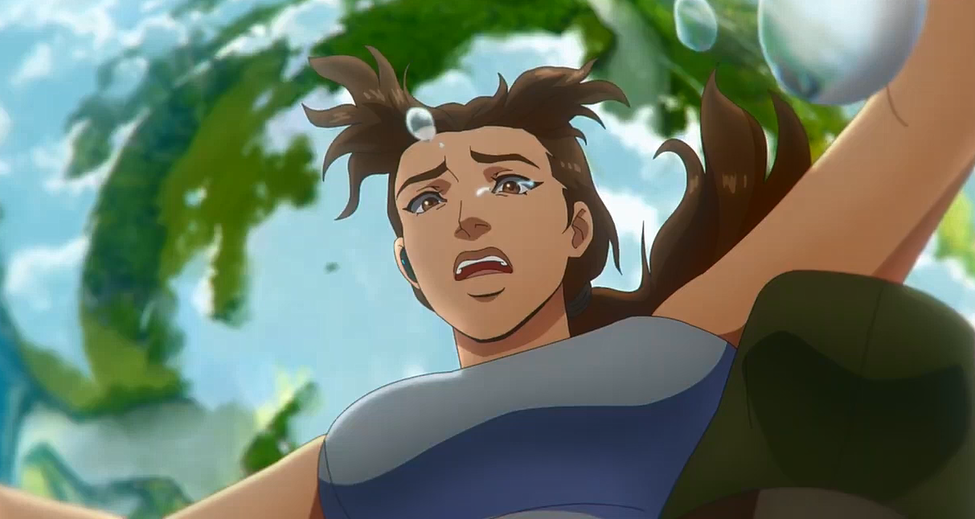}\includegraphics[height=1.5cm,width=0.15\textwidth]{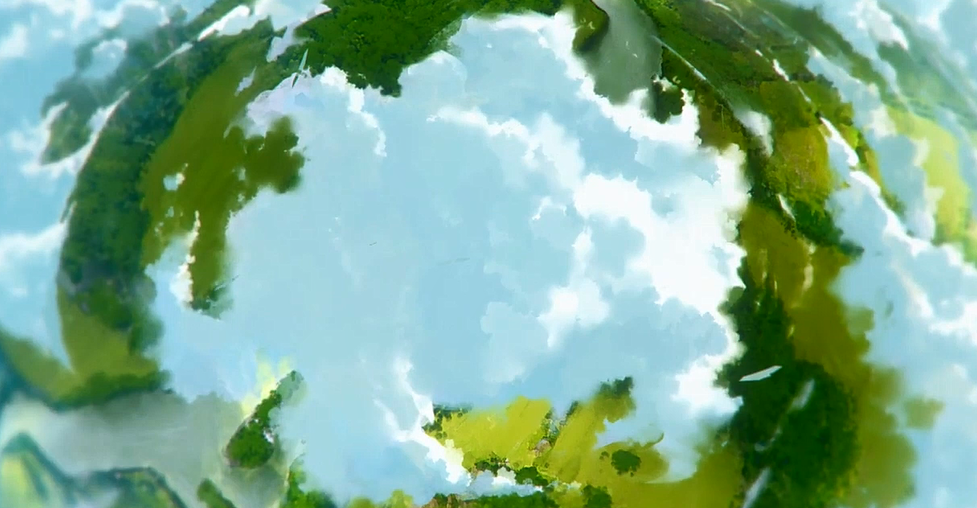} & 37.6623 \\
\includegraphics[height=1.5cm,width=0.15\textwidth]{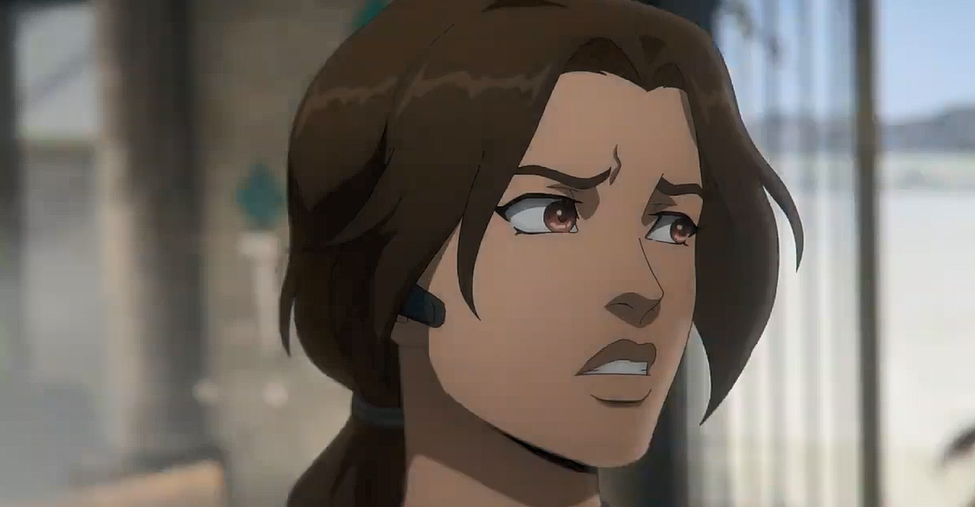}\includegraphics[height=1.5cm,width=0.15\textwidth]{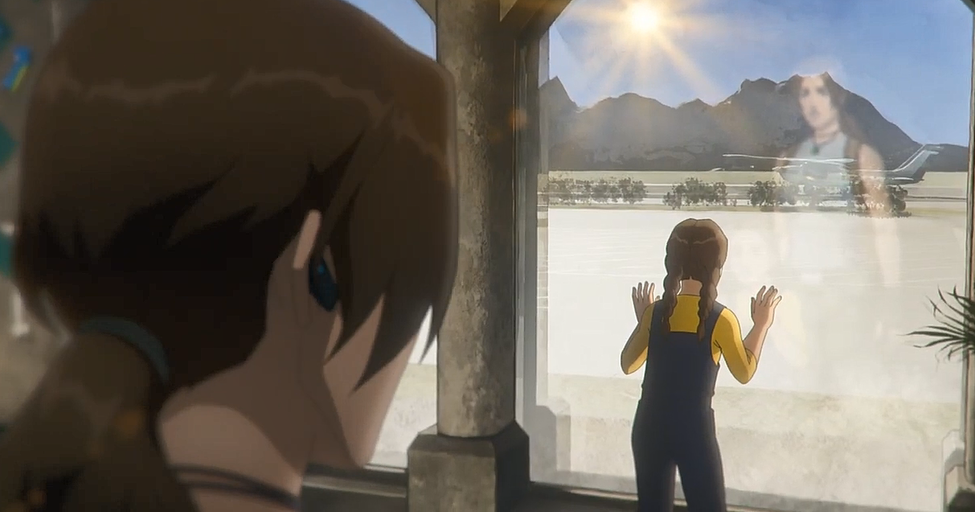} & \textcolor{red}{56.1308} \\
\includegraphics[height=1.5cm,width=0.15\textwidth]{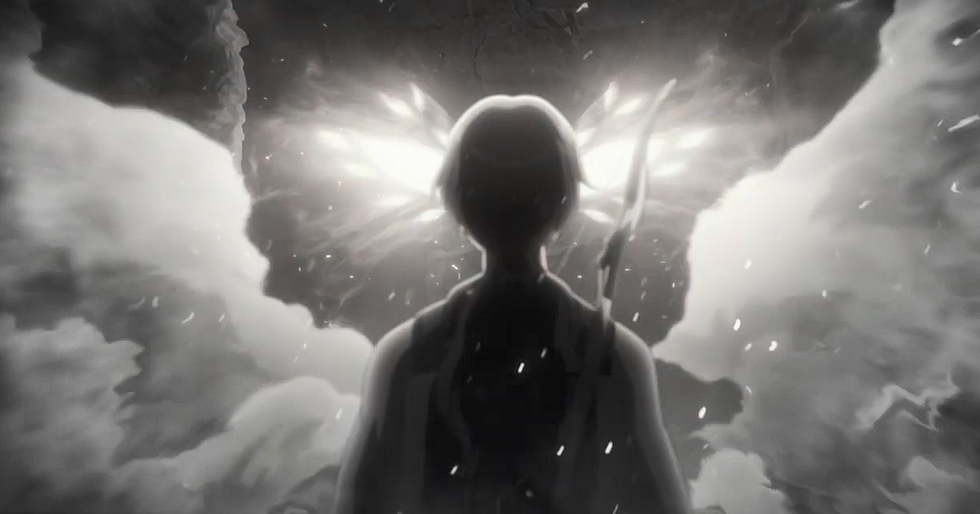}\includegraphics[height=1.5cm,width=0.15\textwidth]{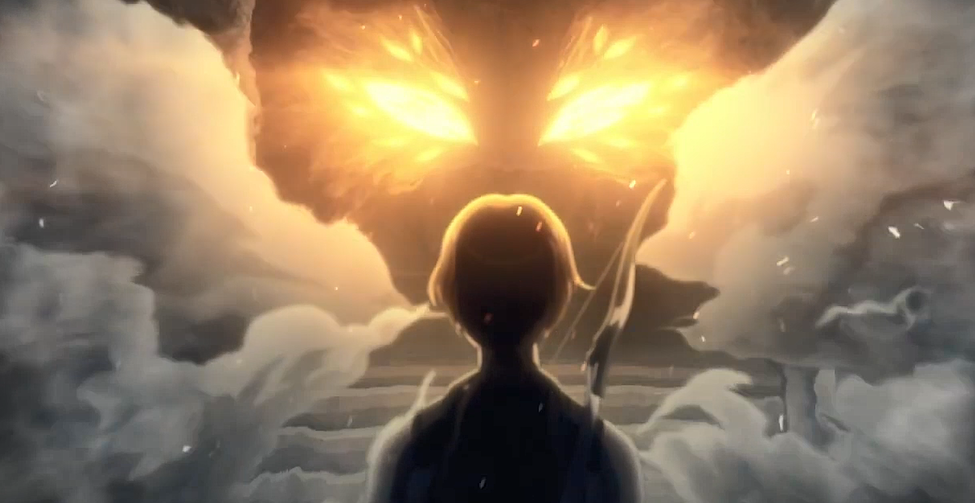} & 42.0404 \\
\bottomrule
\end{tabular}
&
\begin{tabular}{@{}c c@{}}
\toprule
\textbf{Video} & \textbf{MSG Score} \\
\midrule
\includegraphics[height=1.5cm, width=0.15\textwidth]{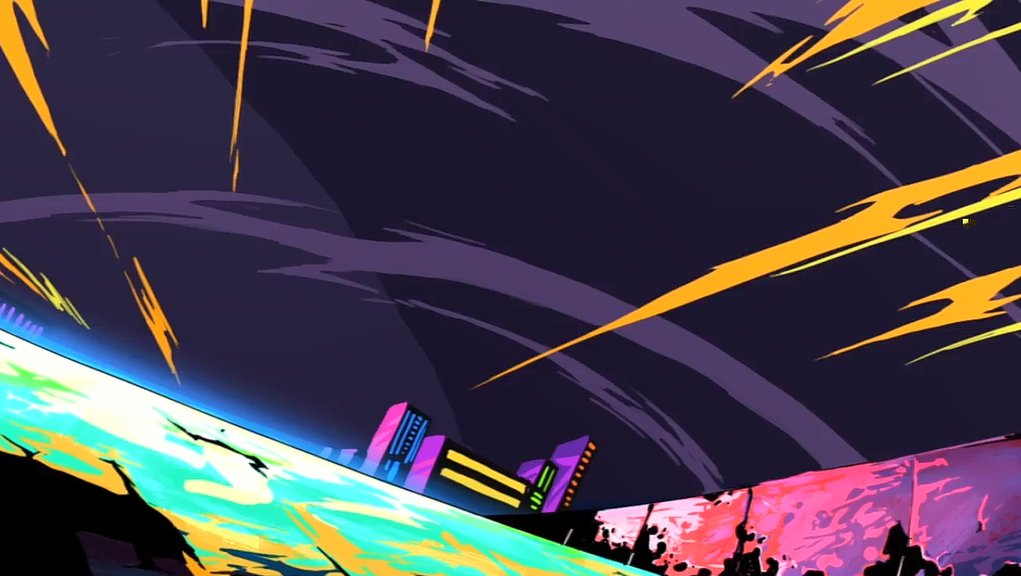}\includegraphics[height=1.5cm, width=0.15\textwidth]{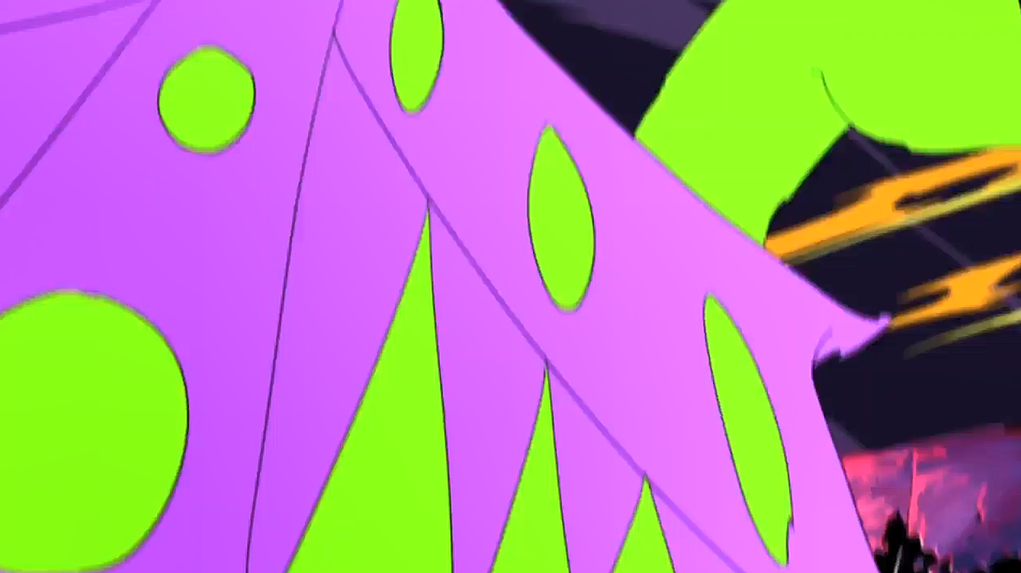} & 25.7409 \\
\includegraphics[height=1.5cm,width=0.15\textwidth]{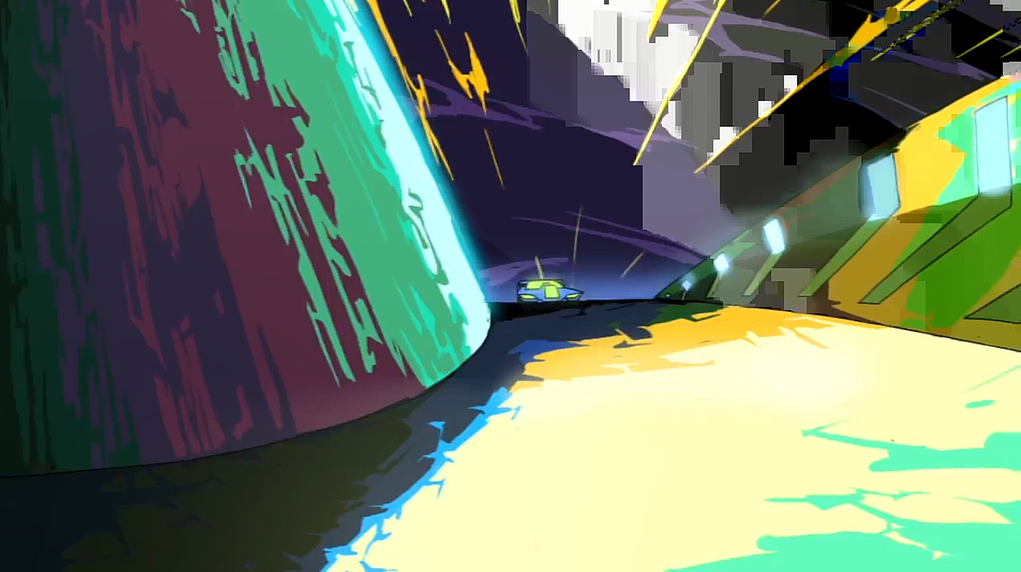}\includegraphics[height=1.5cm, width=0.15\textwidth]{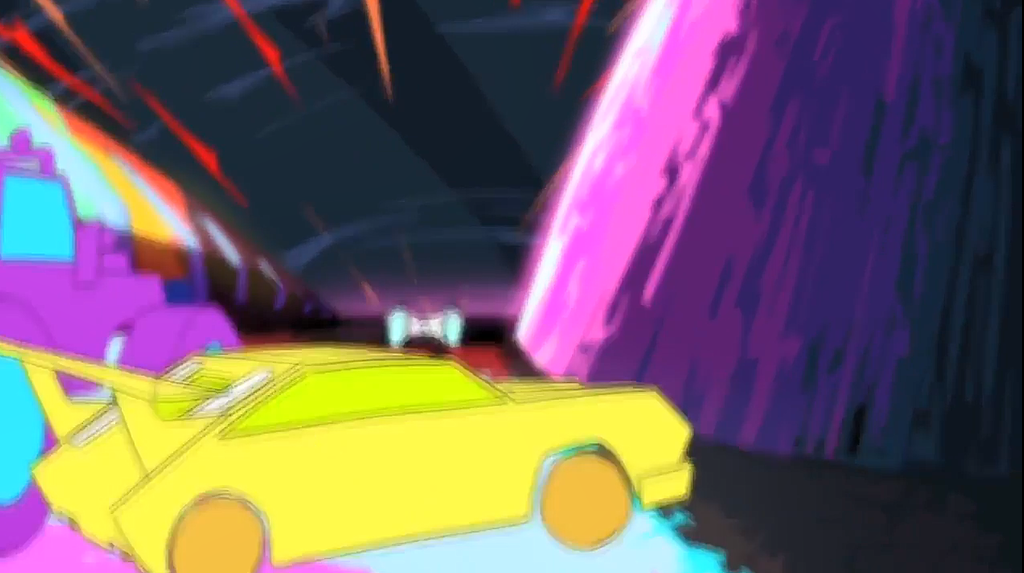} & \textcolor{blue}{22.2168} \\
\includegraphics[height=1.5cm,width=0.15\textwidth]{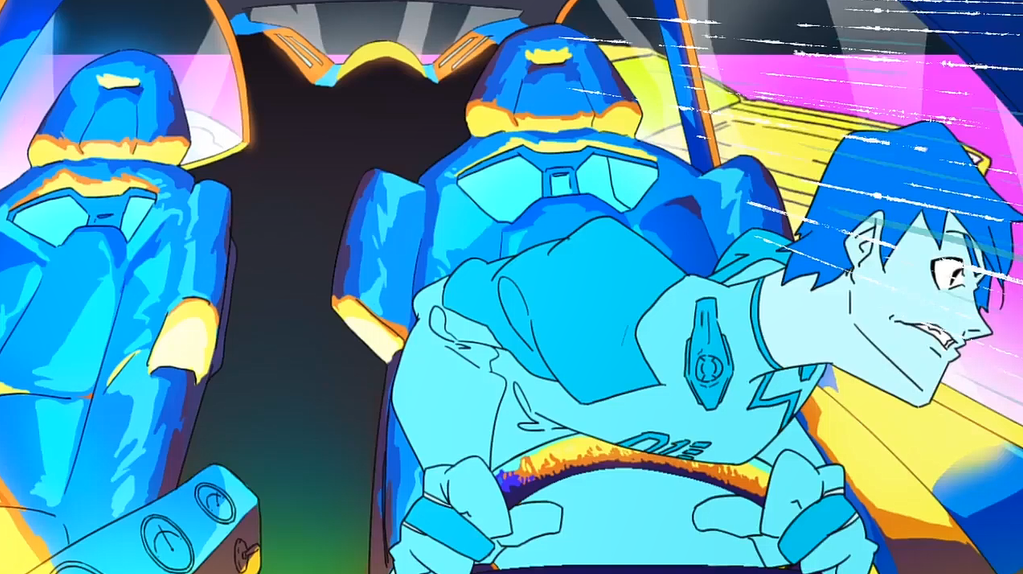}\includegraphics[height=1.5cm, width=0.15\textwidth]{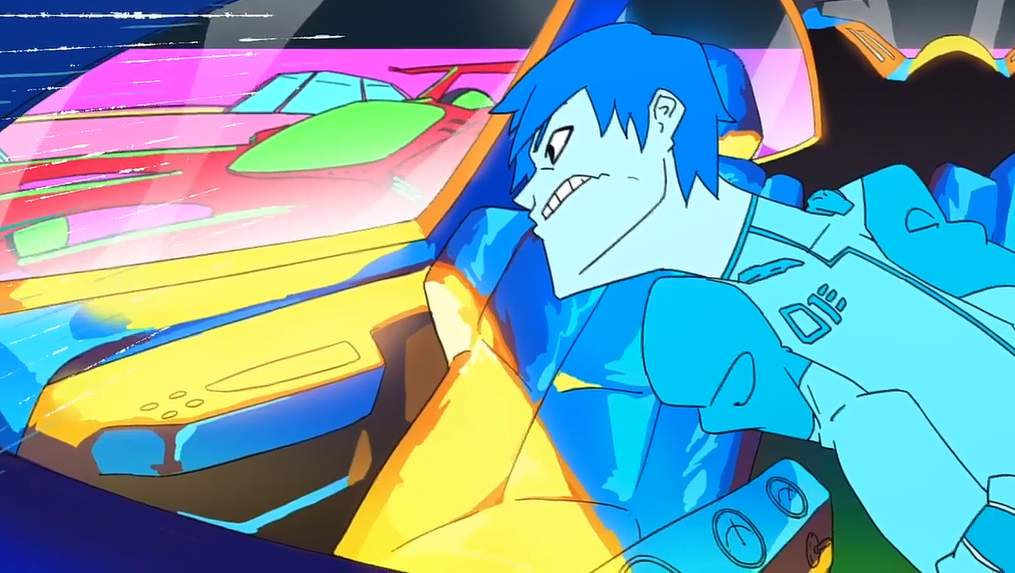} & 46.1107 \\
\includegraphics[height=1.5cm,width=0.15\textwidth]{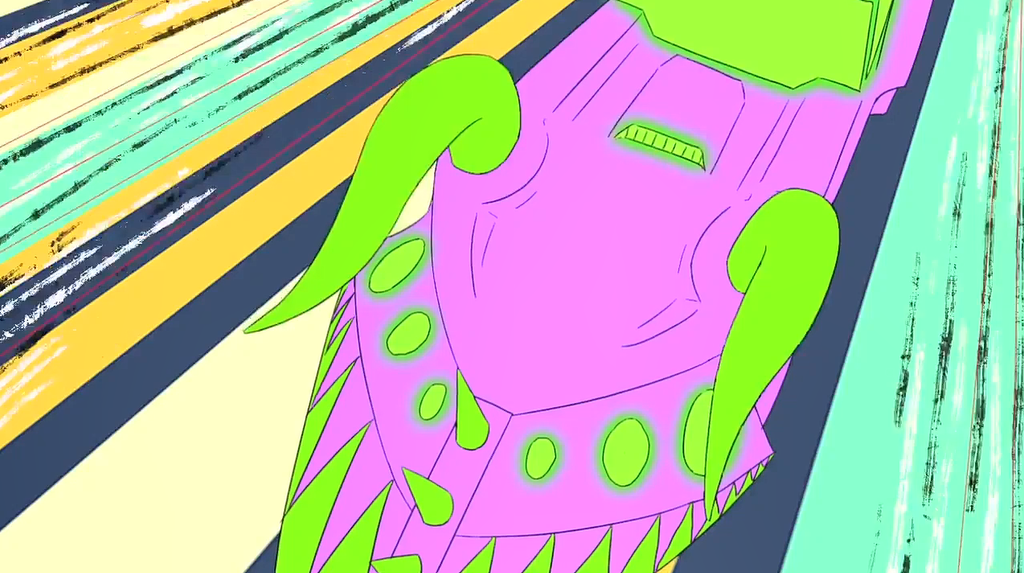}\includegraphics[height=1.5cm, width=0.15\textwidth]{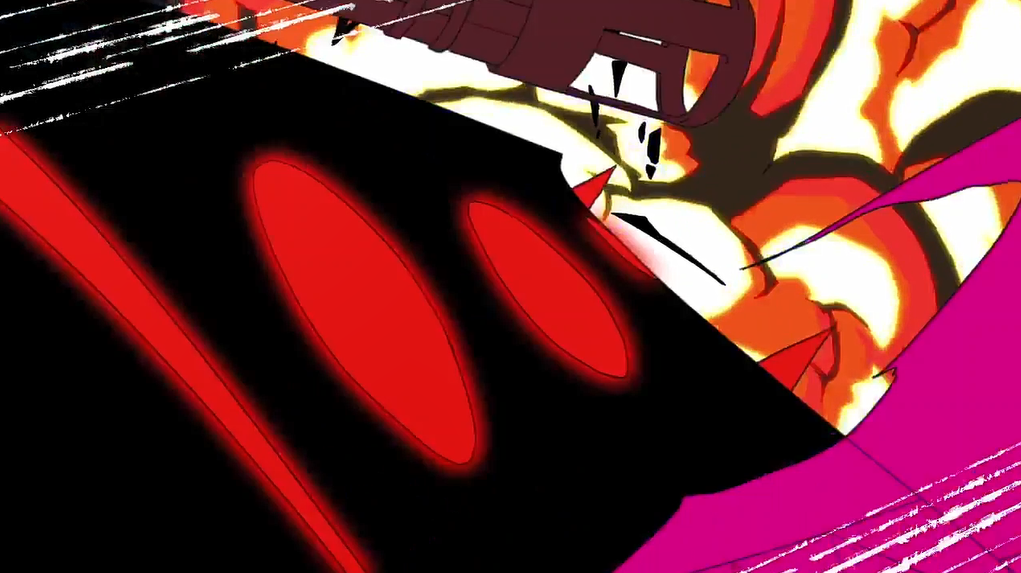} & \textcolor{blue}{18.4185} \\
\includegraphics[height=1.5cm,width=0.15\textwidth]{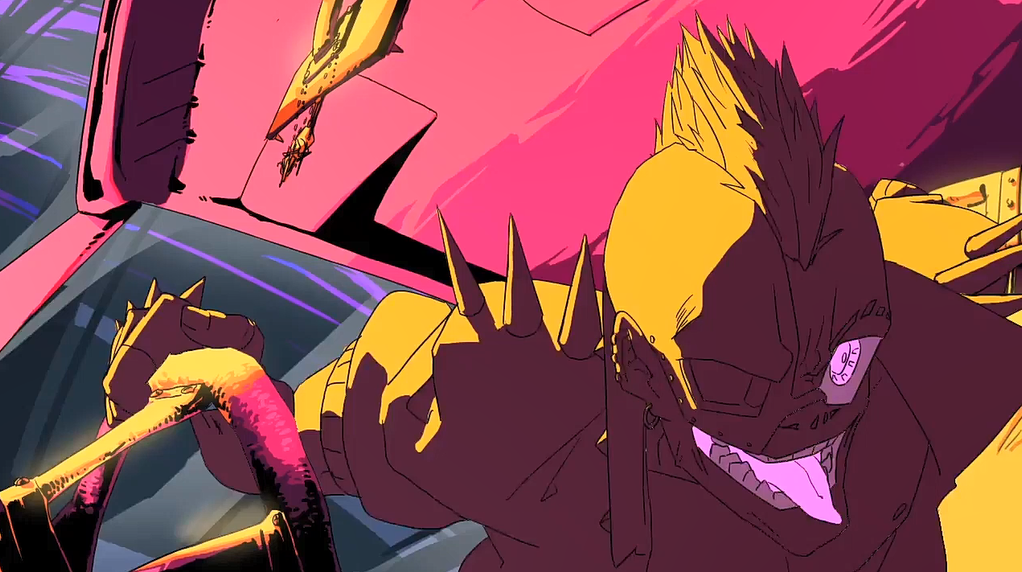}\includegraphics[height=1.5cm, width=0.15\textwidth]{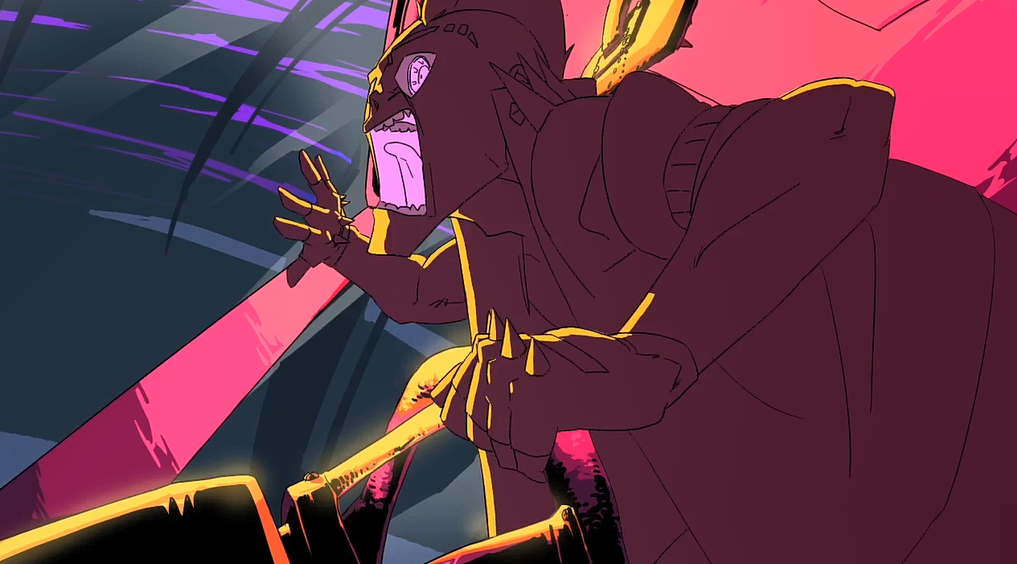} & 43.1217 \\
\includegraphics[height=1.5cm,width=0.15\textwidth]{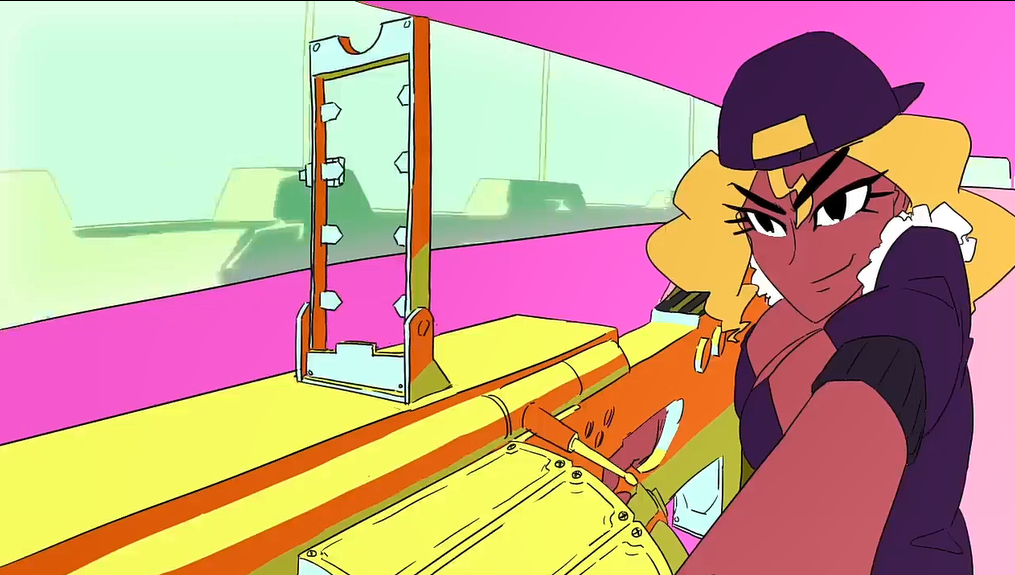}\includegraphics[height=1.5cm, width=0.15\textwidth]{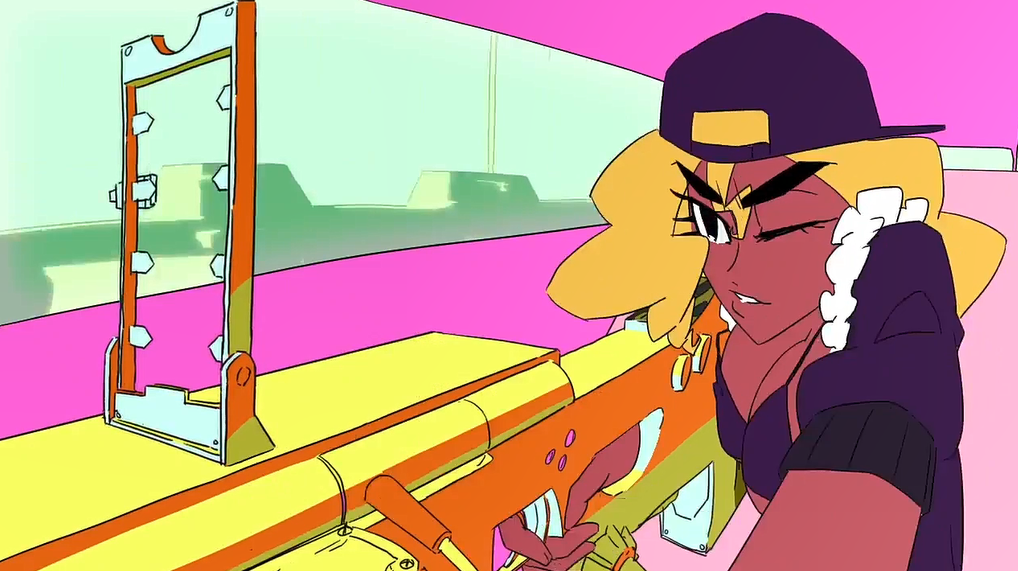} & 40.0531 \\
\bottomrule
\end{tabular} \\
\end{tabular}
\end{table*}

\end{document}